\documentclass[pmlr]{jmlr}

\usepackage{longtable}

\usepackage{booktabs}
\usepackage[load-configurations=version-1]{siunitx} 

\makeatletter
\def\set@curr@file#1{\def\@curr@file{#1}} 
\makeatother

\theorembodyfont{\upshape}
\theoremheaderfont{\scshape}
\theorempostheader{:}
\theoremsep{\newline}

\usepackage{enumitem}
\usepackage{algorithm}
\usepackage{graphics}
\usepackage{algorithmic}
\usepackage{multirow}
\usepackage{adjustbox}

\title[]{Learning by Self-Explanation, with Application to Neural Architecture Search}

\author{\Name{Ramtin Hosseini
}
       \Email{rhossein@eng.ucsd.edu } 
       \AND
       \Name{Pengtao Xie}
       \Email{p1xie@eng.ucsd.edu}\\
       \addr 
University of California San Diego
\AND
       }

\begin{document}

\maketitle

\begin{abstract}
Learning by self-explanation is an effective learning technique in human learning, where students explain a learned topic to themselves for deepening their understanding of this topic. It is interesting to investigate whether this explanation-driven learning methodology broadly used by humans is helpful for improving machine learning as well. Based on this inspiration,  we propose a novel machine learning method called learning by self-explanation (LeaSE). In our approach,  an explainer model improves its learning ability by trying to clearly  explain to an audience model regarding how a prediction outcome is made. LeaSE is formulated as a four-level  optimization problem involving a sequence of four learning stages which are conducted end-to-end in a unified framework: 1) explainer learns; 2) explainer explains; 3) audience learns; 4) explainer re-learns based on the performance of the audience. We develop an efficient algorithm to solve the LeaSE problem.  We apply LeaSE for neural architecture search  on CIFAR-100, CIFAR-10, and ImageNet. Experimental results strongly demonstrate the effectiveness of our method.
\end{abstract}

\section{Introduction}
In humans' learning practice, a broadly adopted learning skill is  self-explanation where students explain to themselves a learned topic to achieve better  understanding of this topic. Self-explanation encourages a student to actively digest and integrate prior knowledge and new information, which helps to fill in the gaps in understanding a topic. It has shown considerable effectiveness in improving learning outcomes.

\begin{figure}[t]
    \centering
 \includegraphics[width=0.5\columnwidth]{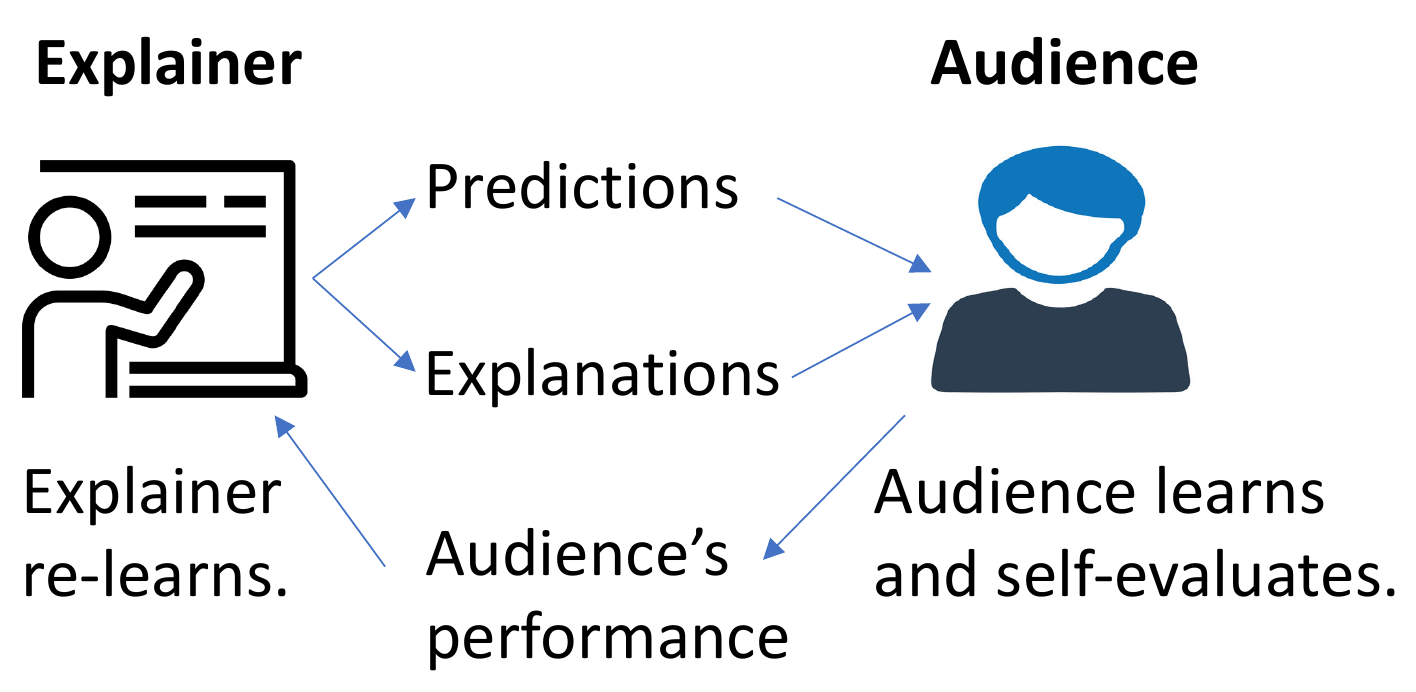}
       \caption{Illustration of learning by self-explanation. An explainer explains to the audience regarding how predictions are made. The audience leverages the explanations to learn and evaluates itself. Performance of the audience acts as feedback to guide the explainer to re-learn for giving better explanations. This process repeats until the audience's performance is good enough.}
 \label{fig:illus}
\end{figure}

We are interested in asking: can the explanation-driven learning method be borrowed from humans to help machines to learn better? Motivated by this inspiration, we propose a novel machine learning framework called learning by self-explanation (LeaSE) (as illustrated in Figure~\ref{fig:illus}). In this framework, there is an explainer model and an audience model. They both learn to undertake the same ML task, such as image classification. The explainer has a learnable architecture and a set of learnable network weights. The audience has a predefined architecture by human experts and a set of learnable network weights. The goal is to help the explainer to learn well on the target task. The way to achieve this goal is to encourage the explainer to give clear explanations to the audience regarding how  predictions are made.  Intuitively, if a model can explain prediction outcomes well, it must have a deep understanding of the prediction task and can learn better based on this understanding. The learning is organized into four stages. In the first stage, the explainer trains its network weights by minimizing the prediction loss on its training dataset, with its architecture fixed. In the second stage, the explainer uses its model trained in the first stage to make predictions on the training data examples of the audience  and leverages an adversarial attack~\citep{goodfellow2014explaining,etmann2019connection} approach to explain the prediction outcomes. In the third stage, the audience model combines its training examples and the explainer-made explanations of prediction outcomes on these examples to train its network weights. In the fourth stage, the explainer updates its neural architecture by minimizing its validation loss and the audience's validation loss.  
The fours stages are synthesized into a unified four-level optimization framework where they are performed jointly in an end-to-end manner. Each learning stage has an influence on other stages. We apply our method for neural architecture search in image  classification tasks. Our method achieves significant improvement on CIFAR-100, CIFAR-10, and ImageNet~\citep{deng2009imagenet}.

The major contributions of this paper are as follows:
\begin{itemize}
\item Drawing inspiration from the explanation-driven learning technique of humans, we propose a novel machine learning approach called learning by self-explanation (LeaSE). In our approach, an explainer model improves its learning ability by trying to clearly explain to an audience model regarding how the prediction outcomes are made. 
\item We develop a multi-level optimization framework to formulate LeaSE which involves four stages of learning: explainer learns; explainer explains; audience learns; explainer re-learns based on the audience's performance. 
\item We develop an efficient algorithm to solve the LeaSE problem. 
\item We apply LeaSE for neural architecture search  on CIFAR-100, CIFAR-10, and ImageNet, where the results demonstrate the effectiveness of our method.
\end{itemize}

\section{Related Works}
\subsection{Neural Architecture Search (NAS)} In the past few years, a wide variety of NAS methods have been proposed and achieved considerable success in automatically identifying highly-performing architectures of neural networks for the sake of reducing the reliance on human experts. Early NAS approaches~\citep{zoph2016neural,pham2018efficient,zoph2018learning} are mostly based on reinforcement learning (RL) which use a policy network to generate architectures and evaluate these architectures on validation set. The validation loss is used as a reward to optimize the policy network and train it to produce high-quality architectures. While RL-based approaches achieve the first wave of success in NAS research, they are computationally very expensive since evaluating the architectures requires a heavy-duty training process. This limitation renders RL-based approaches not applicable for most users who do not have enough computational resources. To address this issue, differentiable search methods \citep{cai2018proxylessnas,liu2018darts,xie2018snas} have been proposed, which parameterize architectures as differentiable functions and perform search using efficient gradient-based methods. In these methods, the search space of architectures is composed of a large set of building blocks where the output of each block is multiplied with a smooth variable indicating how important this block is. Under such a formulation, search becomes solving a mathematical optimization problem defined on the importance variables where the objective is to find out an optimal set of variables that yield the lowest validation loss. This optimization problem can be solved efficiently using gradient-based methods. Differentiable NAS research is initiated by DARTS~\citep{liu2018darts} and further improved by subsequent works such as P-DARTS \citep{chen2019progressive}, PC-DARTS \citep{abs-1907-05737}, etc.  
P-DARTS~\citep{chen2019progressive} grows the depth of architectures progressively in the search process. 
PC-DARTS~\citep{abs-1907-05737} samples sub-architectures from a super network to reduce redundancy during search. 
Our proposed LeaSE framework is orthogonal to existing NAS methods and can be used in combination with any differentiable NAS method to further improve these methods.
\citet{abs-1912-07768} proposed to learn a generative model to generate synthetic examples which are used to search the architecture of an auxiliary model. Our work differs from this one in that: 1) we focus on searching the architecture of a primary model (the explainer) by letting it explain to an auxiliary model (the audience) while \citep{abs-1912-07768} focuses on searching the architecture of the auxiliary model; 2) our primary model produces explanations via adversarial attack while the generative model in \citep{abs-1912-07768} generates synthetic examples. 
Besides RL-based approaches and differentiable NAS, another paradigm of NAS methods~\citep{liu2017hierarchical,real2019regularized} are based on the evolutionary algorithm. In these methods, architectures are formulated as individuals in a population. High-quality architectures produce offspring to replace low-quality architectures, where the quality is measured using fitness scores. Similar to RL-based approaches, these methods also require considerable computing resources.

\subsection{Interpretable Machine Learning}
The explainability and transparency of machine learning models is crucial for mission-critical applications. Many approaches have been proposed for understanding the predictions made by black-box models. Many prior approaches for interpretable ML focus on finding out key evidence from the input data (such as phrases in texts and regions in images) that is most relevant to a prediction, then using these evidence to justify the meaningfulness of the prediction. In \citep{zeiler2014visualizing} and \citep{zintgraf2017visualizing}, the authors perform perturbation  on the input data elements (e.g., pixels) and check which perturbed elements cause more changes of the output. Such elements are considered to be more relevant to the output and are used as explanations. In 
\citep{baehrens2010explain,simonyan2013deep} and \citep{smilkov2017smoothgrad}, the authors identify the contribution of each input feature to the output prediction by propagating the contribution through layers of a deep neural network.  Another body of works~\citep{yang2016hierarchical, mullenbach2018explainable, lei2016rationalizing}  are based on the idea of attention. While training the prediction model, an attention network is trained to calculate an attention score for each input data element. Elements with large attention scores are used as explanations. In \citep{yang2016hierarchical}, attention mechanisms are used to select important words and sentences for interpreting hierarchical recurrent networks. \citet{mullenbach2018explainable} leverage  attention networks for explaining convolution networks.  In addition, another widely adopted idea is to use simple but more interpretable models to interpret expressive black-box models. For example, in LIME~\citep{ribeiro2016should}, an interpretable linear model is used to approximate the decision boundary of a black-box model at an interested instance and interpretation is performed by checking dominant features in the linear model. Our work differs from these existing works in that: existing works focus on interpreting a trained model while our method focuses on improving the training of a model by letting it explain. In other words, the goal of existing works is explaining while that of our work is learning. 
The interpretation module (in the second stage) in our framework can be any model-interpretation method.

The concept of self-explanation was investigated in~\citep{elton2020self}, which calculates confidence levels for  
decisions and explanations based on mutual information. Different from our work, this work does not leverage self-explanation to improve model training. \citet{alvarez2018towards} proposed a self-explaining network (SEN) which simultaneously outputs predictions and explanations. Our work differs from this one in that: our work uses the explanations generated by model A to train model B where B's validation performance reflects how good A's explanations are; the training of A is continuously improved so that it can generate good explanations. Leveraging another model  to evaluate the quality of explanations made by one model is more robust and overfitting-resilient. In contrast, SEN  learns to make predictions and explains  prediction results in a single  model, where the explanations may not be able to generalize in other models.
Explanation-based learning~\citep{dejong1986explanation,minton1990quantitative} has been investigated in logic-based AI systems. These approaches require manual design of logic rules, which are not scalable.

\begin{table}[t]
\caption{Notations in Learning by Self-Explanation}
\centering
\begin{tabular}{l|l}
\hline
Notation & Meaning \\
\hline
$A$ & Architecture of the explainer\\
$E$ & Network weights of the explainer\\
$W$ & Network weights of the audience\\
$\delta$ & Explanations \\
$D_{e}^{(\textrm{tr})}$ & Training data of the explainer\\
$D_{a}^{(\textrm{tr})}$ & Training data of the audience\\
$D_{e}^{(\textrm{val})}$ & Validation data of the explainer\\
$D_{a}^{(\textrm{val})}$ & Validation data of the audience\\
\hline
\end{tabular}
\label{tb:notations}
\end{table}

\begin{figure}[t]
    \centering
 \includegraphics[width=0.5\columnwidth]{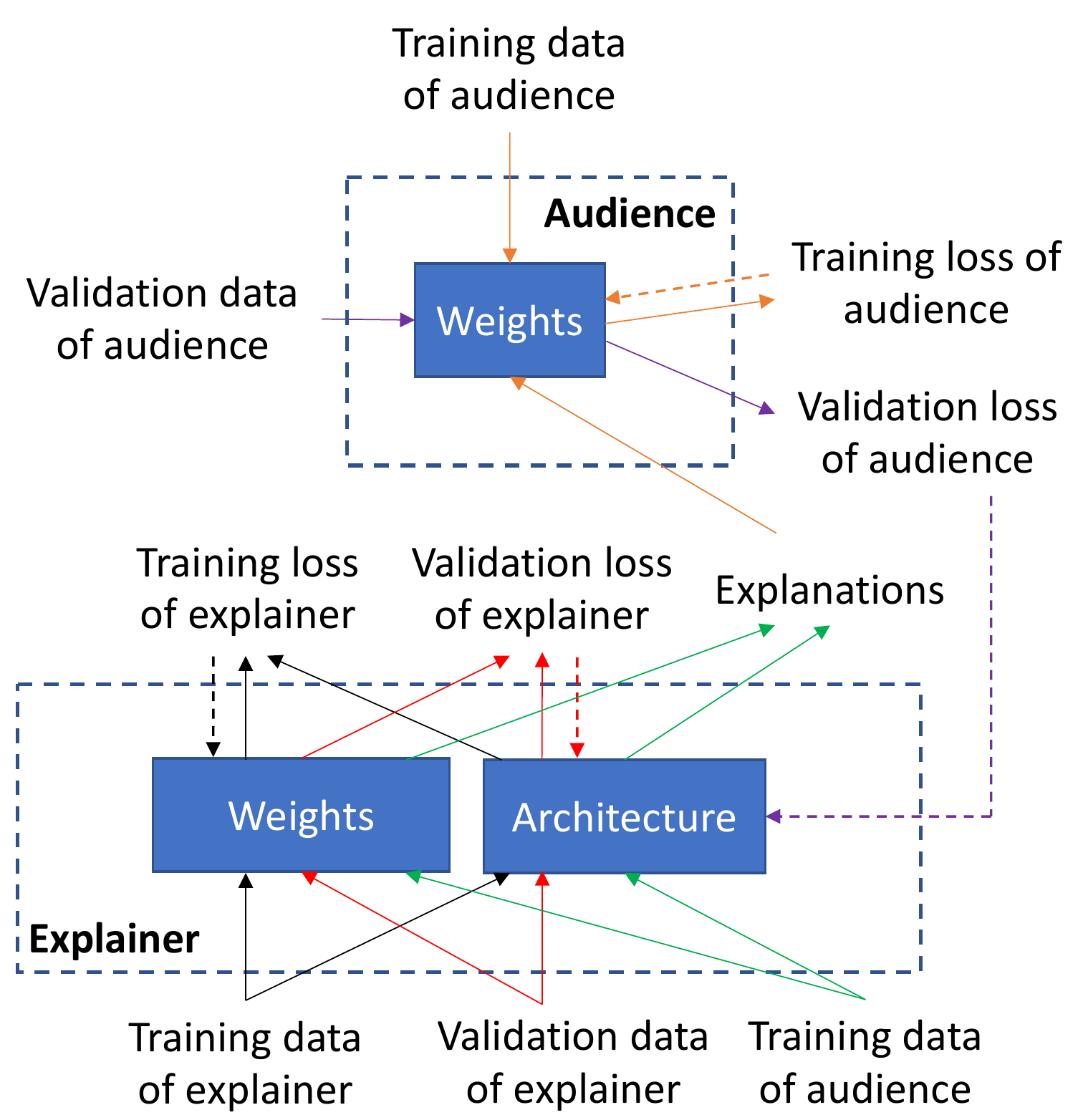}
       \caption{Learning by self-explanation. Following the solid arrows, predictions are made and training/validation losses are calculated. Following the dotted arrows, gradient updates of network weights and architecture variables are calculated and weights/architecture are updated.
       }
 \label{fig:arch}
\end{figure}

\section{Methods}
In this section, we propose a four-level optimization framework to formulate learning by self-explanation (LeaSE) (as shown in Figure~\ref{fig:arch}) and develop an optimization algorithm to solve the four-level optimization problem.

\subsection{Learning by Self-Explanation}
In the LeaSE framework, there is an explainer model and an audience model, both of which learn to perform the same target task. The primary goal of our framework is to help the explainer to learn the target task very well. The way to achieve this goal is to let the explainer make meaningful explanations of the prediction outcomes in the target task. The intuition behind LeaSE is: to correctly explain prediction results, a model needs to learn to understand the target task very well. The explainer has a learnable architecture $A$ and a set of learnable network weights $E$. The audience has a pre-defined neural architecture (by human experts) and a set of learnable network weights $W$. The learning is organized into four stages. In the first stage, the explainer trains its network weights $E$ on its training dataset $D_e^{(\textrm{tr})}$, with the architecture fixed:
\begin{equation}
    E^*(A) =\textrm{min}_{E} \; L(E,A,D_e^{(\textrm{tr})}). 
\end{equation}
To define the training loss, we need to use the architecture $A$ together with network weights $W$ to make predictions on training examples. However, $A$ cannot be updated by minimizing the training loss. Otherwise,  a trivial solution of $A$ will be yielded: $A$ is very large and complex that it can perfectly overfit the training data but  will make largely incorrect predictions on novel data examples.   Note that  $E^*(A)$ is a function of $A$ for that $L(E,A, D_e^{(\textrm{tr})})$ is a function of $A$ and $E^*(A)$ depends on $L(E, A,D_e^{(\textrm{tr})})$.  In the second stage, the explainer uses the trained model $E^*(A)$ to make predictions on the input training examples $D_a^{(\textrm{tr})}$ of the audience and explains the prediction outcomes. Specifically, given an input data example $x$ (without loss of generality, we assume it is an image) and the predicted label $y$, the explainer aims to find out a subset of image patches $P$ in $x$ that are mostly correlated with $y$ and uses $P$ as explanations for $y$. We leverage an adversarial attack approach~\citep{goodfellow2014explaining,etmann2019connection} to achieve this goal. Adversarial attack adds small random perturbations $\delta$ to pixels in $x$ so that the prediction outcome on the perturbed image $x+\delta$ is no longer $y$. Pixels that are perturbed more have higher correlations with the prediction outcome $y$ and can be used as explanations. This process amounts to solving the following optimization problem:
\begin{equation}
   \Delta^*(E^*(A)) =\textrm{max}_{\Delta} \;\; \sum_{i=1}^N \ell(f(x_i+\delta_i;{E^*(A)}),f(x_i;{E^*(A)})) 
   \label{eq:explain}
\end{equation}
where $\Delta=\{\delta_i\}_{i=1}^N$ and $\delta_i$ is the perturbation added to image $x_i$. $f(x_i+\delta_i;{E^*(A)})$ and $f(x_i;{E^*(A)})$ are the prediction outcomes of the explainer's network $f(\cdot;{E^*(A)})$ on $x_i+\delta_i$ and $x_i$. Without loss of generality, we assume the task is image classification (with $K$ classes). Then $f(x_i+\delta_i;{E^*(A)})$ and $f(x_i;{E^*(A)})$ are both $K$-dimensional vectors containing prediction probabilities on the $K$ classes. $\ell(\cdot,\cdot)$ is the cross-entropy loss with $\ell(\mathbf{a},\mathbf{b})=-\sum_{k=1}^K b_i\log a_i$. In this optimization problem, the explainer aims to find out perturbations for each image so that the predicted outcome on the perturbed image is largely different from that on the original image. The learned optimal perturbations are used as explanations and those with larger values indicate that the corresponding pixels are more important in decision-making.  $\Delta^*(E^*(A))$ is a function of $E^*(A)$ since $\Delta^*(E^*(A))$ is a function of the objective in Eq.(\ref{eq:explain}) and the objective is a function of $E^*(A)$. In the third stage, given the explanations $\Delta^*(E^*(A))$ made by the explainer, the audience leverages them to learn the target task. Since the perturbations indicate how important the input pixels are, the audience uses them to reweigh the pixels: $x\odot \delta$, where $\odot$ denotes element-wise multiplication. Pixels that are more important are given more weights. Then the audience trains its network weights on these weighted images:
\begin{equation}
      W^*( \Delta^*(E^*(A)) ) =\textrm{min}_{W} \;\; \sum_{i=1}^N \ell(f(\delta_i^*(E^*(A))\odot x_i;W),t_i),
      \label{eq:w}
\end{equation}
where $f(\delta_i^*(E^*(A))\odot x_i;W)$ is the prediction outcome of the audience's network $f(\cdot;W)$ on the weighted image $\delta_i^*(E^*(A))\odot x_i$ and $t_i$ is the class label.  $W^*( \Delta^*(E^*(A)) )$ is a function of $\Delta^*(E^*(A))$ since $W^*( \Delta^*(E^*(A)) )$ is a function of the objective in Eq.(\ref{eq:w}) and the objective is a function of  $\Delta^*(E^*(A))$.  In the fourth stage, the explainer validates its network weights $E^*(A)$ on its validation set $D^{(\textrm{val})}_e$ and the audience validates its network weights $W^*( \Delta^*(E^*(A)) )$ on its validation set $D^{(\textrm{val})}_a$. The explainer optimizes its architecture by minimizing its validation loss and the audience's validation loss:
\begin{equation}
    \textrm{min}_{A}  
    \; L(E^*(A),A, D_e^{(\textrm{val})})  + \gamma L(W^*( \Delta^*(E^*(A)) ), D_a^{(\textrm{val})}),
\end{equation}
where $\gamma$ is a tradeoff parameter. 

We integrate the four stages in a unified four-level optimization framework and obtain the following formulation of LeaSE: 
\begin{equation}
\begin{array}{l}
\underset{A}{\textrm{min}}
  \;\;  
    \; L(E^*(A),A, D_e^{(\textrm{val})})  + \gamma L(W^*( \Delta^*(E^*(A)) ), D_a^{(\textrm{val})})\\
      s.t. \;\;\; W^*( \Delta^*(E^*(A)) ) =
      \underset{W}{\textrm{min}}
      \;\; \sum\limits_{i=1}^N \ell(f(\delta_i^*(E^*(A))\odot x_i;W),t_i)\\
    \quad\quad \Delta^*(E^*(A)) =
    \underset{\Delta}{\textrm{max}}
    \;\; \sum\limits_{i=1}^N \ell(f(x_i+\delta_i;{E^*(A)}),f(x_i;{E^*(A)})) \\
     \quad\quad   E^*(A) =
     \underset{E}{\textrm{min}}
     \; L(E,A,D_e^{(\textrm{tr})}). 
\end{array}
\label{eq:lease}
\end{equation}
In this framework, there are four optimization problems, each corresponding to a learning stage. From bottom to up, the optimization problems correspond to learning stage 1, 2, 3, and 4 respectively. The first three optimization problems are nested on the constraint of the fourth optimization problem. These four stages are conducted end-to-end in this unified framework. The solution $E^*(A)$ obtained in the first stage is used to perform explanation in the second stage. The explanations $\Delta^*(E^*(A))$ obtained in the second stage are used to train the model in the third stage. The solutions obtained in the first and third stage are used to make predictions on the fourth stage. The  architecture $A$ updated in the fourth stage changes the training loss in the first stage and consequently changes the solution $E^*(A)$, which subsequently changes $\Delta^*(E^*(A))$ and $W^*( \Delta^*(E^*(A)) )$. Following \citep{liu2018darts}, we perform  differentiable search on $A$ in a search space composed of candidate building blocks. Searching amounts to selecting a subset of candidate blocks by learning a selection  variable for each block. The selection variables indicate the importance of individual blocks and are differentiable.

\section{ Optimization Algorithm}
\label{sec:alg}
We develop an efficient algorithm to solve the LeaSE problem. Getting insights from \citep{liu2018darts},
we calculate the gradient of $L(E,A, D_e^{(\textrm{tr})})$ w.r.t $E$ and approximate $E^{*}(A)$ using one-step gradient descent update of $E$.
We plug  the approximation $E'$ of $E^{*}(A)$ into $\sum_{i=1}^N \ell(f(x_i+\delta_i;{E^*(A)}),f(x_i;{E^*(A)}))$ and obtain an approximated objective denoted by $O_{\Delta}$. Then we approximate  $\Delta^*(E^*(A))$ using one-step gradient descent update of  $\Delta$ based on the gradient of  $O_{\Delta}$. Next, we plug  the approximation $\Delta'$ of $\Delta^*(E^*(A))$ into $\sum_{i=1}^N \ell(f(\delta_i^*(E^*(A))\odot x_i;W),t_i)$  and get another approximated objective denoted by $O_W$. Then we approximate  $W^*( \Delta^*(E^*(A)) )$ using one-step gradient descent update of  $W$ based on the gradient of $O_{W}$. Finally, we plug  the approximation $E'$ of $E^*(A)$ and the approximation $W'$ of $W^*( \Delta^*(E^*(A)) )$ into $L(E^*(A),A, D_e^{(\textrm{val})})  + \gamma L(W^*( \Delta^*(E^*(A)) ), D_a^{(\textrm{val})})$  and get the third approximated objective denoted by $O_A$. 
$A$ is updated by descending the gradient of $O_A$. 
In the sequel, we use $\nabla^2_{Y,X}f(X,Y)$ to denote $\frac{\partial f(X,Y)}{\partial X\partial Y}$.

First of all, we approximate $E^{*}(A)$ using 
\begin{equation}
    E'=E - \xi_{e}  \nabla_{E}L(E, A, D_{e}^{(\mathrm{tr})})
    \label{eq:update_e}
\end{equation}
where $\xi_{e}$ is a learning rate. Plugging  $E'$ into $\sum_{i=1}^N \ell(f(x_i+\delta_i;{E^*(A)}),f(x_i;{E^*(A)}))$, we obtain an approximated objective $O_{\Delta}=\sum_{i=1}^N \ell(f(x_i+\delta_i;E'),f(x_i;E'))$. Then  we approximate  $\Delta^*(E^*(A))$ using one-step gradient descent update of  $\Delta$ with respect to $O_{\Delta}$:
\begin{equation}
    \Delta'=\Delta - \xi_{\Delta}  \nabla_{\Delta}(\sum\limits_{i=1}^N \ell(f(x_i+\delta_i;E'),f(x_i;E'))).
    \label{eq:update_delta}
\end{equation}
Plugging $\Delta'$ into $\sum_{i=1}^N \ell(f(\delta_i^*(E^*(A))\odot x_i;W),t_i)$, we obtain an approximated objective $O_{W}=\sum_{i=1}^N \ell(f(\delta'_i\odot x_i;W),t_i)$.
 Then  we approximate  $W^*( \Delta^*(E^*(A)) )$ using one-step gradient descent update of  $W$ with respect to $O_{W}$:
\begin{equation}
    W'=W - \xi_{W}  \nabla_{w}(\sum\limits_{i=1}^N \ell(f(\delta'_i\odot x_i;W),t_i)).
    \label{eq:update_w}
\end{equation}
Finally, we plug  $E'$ and $W'$ into $ L(E^*(A), D_e^{(\textrm{val})})  + \gamma L(W^*( \Delta^*(E^*(A)) ), D_a^{(\textrm{val})})$ and get $O_A=L(E', D_e^{(\textrm{val})})  + \gamma L(W', D_a^{(\textrm{val})})$. We can update the explainer's architecture $A$ by descending the gradient of $O_A$ w.r.t $A$:
\begin{equation}
\begin{array}{l}
A\gets A-\eta ( \nabla_A L(E', D_e^{(\textrm{val})}) +\gamma \nabla_A L(W', D_a^{(\textrm{val})}))   
    \end{array}
    \label{eq:update_a}
\end{equation}
where 
\begin{equation}
\begin{array}{l}
     \nabla_{A} L(E',A,  D_e^{(\textrm{val})}) =  \\
     \nabla_{A}  L(E - \xi_{e}  \nabla_{E}L(E, A, D_{e}^{(\mathrm{tr})}),A,  D_e^{(\textrm{val})})=\\
     - \xi_{e}  \nabla^2_{A,E}L(E, A, D_{e}^{(\mathrm{tr})})\nabla_{E'} L(E',A,  D_e^{(\textrm{val})})+ \nabla_{A} L(E', A, D_e^{(\textrm{val})})
\end{array}
\label{eq:descent_arch}
\end{equation}
The first term in the third line   involves expensive matrix-vector product, whose computational complexity can be reduced by a finite difference approximation:
\begin{equation}
\begin{array}{ll}
     \nabla_{A, E}^{2} L(E,A, D_{e}^{(\mathrm{tr})})\nabla_{E'} L(E',A,D_e^{(\textrm{val})})\approx
     \frac{1}{2\alpha}(\nabla_{A} L( E^{+},A, D_{e}^{(\mathrm{tr})})-\nabla_{A} L(E^{-},A, D_{e}^{(\mathrm{tr})})),
\end{array}
\label{eq:finite-aw}
\end{equation}
where $E^{\pm}=E \pm \alpha \nabla_{E^{\prime}} L(E',A, D_e^{(\textrm{val})})$ and $\alpha$ is a small scalar that equals $0.01 /\|\nabla_{E'} L(E',A,$\\$D_e^{(\textrm{val})})\|_{2}$.

For $\nabla_A L(W', D_a^{(\textrm{val})})$ in Eq.(\ref{eq:update_a}), it can be calculated as $\frac{\partial E'}{\partial A}
\frac{\partial \Delta'}{\partial E'}
\frac{\partial W'}{\partial \Delta'}
\nabla_{W'}L(W',D_a^{(\textrm{val})})$ according to the chain rule, where 
\begin{align}
\frac{\partial W'}{\partial \Delta'}&=\frac{\partial (W - \xi_{w}  \nabla_{W}(\sum_{i=1}^N \ell(f(\delta'_i\odot x_i;W),t_i)))}{\partial \Delta'}\\
& = - \xi_{W} \nabla^2_{\Delta',W}(\sum_{i=1}^N \ell(f(\delta'_i\odot x_i;W),t_i)),
\end{align}
\begin{align}
\frac{\partial \Delta'}{\partial E'}&=\frac{\partial (\Delta - \xi_{\Delta}  \nabla_{\Delta}(\sum\limits_{i=1}^N \ell(f(x_i+\delta_i;E'),f(x_i;E'))))}{\partial E'}\\
& = - \xi_{\Delta} \nabla^2_{E',\Delta}(\sum\limits_{i=1}^N \ell(f(x_i+\delta_i;E'),f(x_i;E'))),
\end{align}
and 
\begin{align}
    \frac{\partial E' }{\partial A}&=\frac{\partial (E - \xi_{e}  \nabla_{E}L(E, A, D_{e}^{(\mathrm{tr})}))}{\partial A}\\
     & = - \xi_{e}   \nabla^2_{A,E}L(E, A, D_{e}^{(\mathrm{tr})}).
\end{align}

This algorithm is summarized in Algorithm~\ref{algo:algo}.
\begin{algorithm}[H]
\SetAlgoLined
 \While{not converged}{
1. Update the explainer's  weights $E$ using Eq.(\ref{eq:update_e})
\\
2. Update the explanations $\Delta$ using Eq.(\ref{eq:update_delta})
\\
2. Update the audience's  weights $W$ using Eq.(\ref{eq:update_w})
\\
3. Update the explainer's architecture $A$ using Eq.(\ref{eq:update_a})
 }
 \caption{Optimization algorithm for learning by self-explanation}
 \label{algo:algo}
\end{algorithm}

\section{Experiments}
In the experiments, we apply our proposed LeaSE framework to perform neural architecture search for image classification. Each experiment consists of two phrases: architecture search and evaluation. In the search phrase, an optimal cell is identified. In the evaluation phrase, multiple copies of the optimal cell are stacked into a larger network, which is retrained from scratch.

\subsection{Datasets}
The experiments are performed on three datasets, including CIFAR-10, CIFAR-100, and ImageNet~\citep{deng2009imagenet}. Both CIFAR-10 and CIFAR-100 contain 60K images from 10 classes (each class has the same number of images). For each of them, we split it into a training set with 25K images, a validation set with 25K images, and a test set with 10K images. During architecture search in LeaSE, the training set is used as $D_{e}^{(\textrm{tr})}$ and $D_{a}^{(\textrm{tr})}$ and the validation set is used as $D_{e}^{(\textrm{val})}$ and $D_{a}^{(\textrm{val})}$. During architecture evaluation, the composed large network is trained on the combination of $D_{e}^{(\textrm{tr})}$ and $D_{a}^{(\textrm{tr})}$.  ImageNet contains 1.2M training images and 50K test images, coming from 1000 objective classes. Performing architecture search on the 1.2M images is computationally too costly. To address this issue, following~\citep{abs-1907-05737}, we randomly sample 10\% images from the 1.2M images to form a new training set and another 2.5\% images to form a validation set, then perform search on them. During architecture evaluation, the composed large network is trained on the entire set of 1.2M images.

\subsection{Experimental Settings}
Our framework is orthogonal to existing NAS approaches and can be applied to any differentiable NAS method. In the experiments, LeaSE was applied to DARTS~\citep{liu2018darts}, P-DARTS~\citep{chen2019progressive}, and PC-DARTS~\citep{abs-1907-05737}. The search spaces of these methods are composed of 
(dilated) separable convolutions with sizes of $3\times 3$ and $5\times 5$, max pooling with size of $3\times 3$, average pooling with size of $3\times 3$, identity, and zero. Each LeaSE experiment was repeated for ten times with different random seeds. The mean and standard deviation of classification errors obtained from the 10 runs are reported.

During architecture search, for  CIFAR-10 and CIFAR-100, the architecture of the explainer is a stack of 8 cells.
Each cell consists of 7 nodes.  We set the initial channel number to 16.  For the architecture of the audience model, we experimented with ResNet-18 and ResNet-50~\citep{resnet}. We set the tradeoff parameter $\gamma$ to 1. The search algorithm was based on SGD, with a batch size of 64, an initial learning rate of 0.025 (reduced in later epochs using a cosine decay scheduler), an epoch number of 50, a weight decay of 3e-4, and a momentum of 0.9. The rest of hyperparameters mostly follow those in DARTS, P-DARTS, and PC-DARTS.

During architecture evaluation, for CIFAR-10 and CIFAR-100, a larger network of the explainer is formed by stacking 20 copies of the searched cell. The initial channel number was set to 36. We trained the network with a batch size of 96, an epoch number of 600, on a single  Tesla v100 GPU.  On ImageNet, we evaluate two types of architectures: 1) those searched on a subset of  ImageNet; 2) those searched on CIFAR-10 or CIFAR-100. In either type, 14 copies of optimally searched cells  are stacked into a large network, which was trained using eight Tesla v100 GPUs on the 1.2M training images, with a batch size of 1024 and an epoch number of 250. Initial channel number was set to 48.

\subsection{Results}

\begin{table}[t]
\caption{Test error on CIFAR-100, number of model weights (millions), and search cost (GPU days on a Tesla v100).  
    DARTS-1st  and 
DARTS-2nd represents that first-order and second-order approximation is used in DARTS' optimization algorithm. LeaSE-R18-DARTS1st represents that the manually-designed architecture in the audience model is ResNet-18 and the search space is the same as that in DARTS-1st. Similar meanings hold for other notations in such a format. R50 denotes ResNet-50. 
    Results marked with * are obtained from DARTS$^{-}$ \citep{abs-2009-01027}. Methods marked with $\dag$ were re-run for 10 times. For DARTS$^{+}$ marked with $\Delta$, we ran it for 600 epochs instead of 2000 epochs (used in \citep{abs-1909-06035}) in the architecture evaluation stage, to ensure the comparison with other methods (running 600 epochs) is fair. 
    }
    \centering
    \begin{tabular}{l|ccc}
    \toprule
    Method & Error(\%)& Param(M)& Cost\\
    \midrule
    *ResNet \citep{he2016deep}&22.10&1.7&-\\
     *DenseNet \citep{HuangLMW17}&17.18&25.6 &-\\
    \hline
    *PNAS \citep{LiuZNSHLFYHM18}&19.53&3.2&150\\
    *ENAS \citep{pham2018efficient}&19.43&4.6&0.5\\
        *AmoebaNet \citep{real2019regularized}&18.93&3.1&3150\\
    \hline
    *GDAS \citep{DongY19}&18.38&3.4&0.2\\
    *R-DARTS \citep{ZelaESMBH20}&18.01$\pm$0.26&-&1.6
    \\
       *DARTS$^{-}$ \citep{abs-2009-01027}&17.51$\pm$0.25&3.3&0.4\\
      ${}^{\dag}$DARTS$^{-}$ \citep{abs-2009-01027}& 18.97$\pm$0.16& 3.1&0.4\\
          ${}^{\Delta}$DARTS$^{+}$ \citep{abs-1909-06035}&17.11$\pm$0.43&3.8&0.2\\
      *DropNAS \citep{HongL0TWL020} & 16.39&4.4&0.7 \\
\hline
\hline
     ${}^{\dag}$DARTS-1st \citep{liu2018darts}  &20.52$\pm$0.31 &3.5 &1.0\\
     $\;\;$LeaSE-RN18-DARTS1st (ours)  &17.04$\pm$0.10&3.6 & 1.1 \\
     $\;\;$LeaSE-RN50-DARTS1st (ours) &\textbf{16.87}$\pm$0.08 &3.6 &1.2 \\
     \hline
            *DARTS-2nd \citep{liu2018darts}  & 20.58$\pm$0.44&3.5&1.5 \\
             $\;\;$LeaSE-RN18-DARTS2nd (ours) &16.80$\pm$0.17 &3.7 & 1.7\\
               $\;\;$LeaSE-RN50-DARTS2nd (ours) & \textbf{16.39}$\pm$0.07  & 3.6& 1.9\\
            \hline
     
           $\dag$PC-DARTS \citep{abs-1907-05737} &17.96$\pm$0.15&3.9&0.1\\
       $\;\;$LeaSE-RN18-PCDARTS (ours) &16.39$\pm$0.21 &4.0 &0.5  \\
              $\;\;$LeaSE-RN50-PCDARTS (ours)& \textbf{16.17}$\pm$0.05 &3.9 & 0.7  \\
              \hline
                    *P-DARTS \citep{chen2019progressive}&17.49&3.6&0.3\\
       $\;\;$LeaSE-RN18-PDARTS (ours) &15.23$\pm$0.11 &3.7 &0.8 \\
              $\;\;$LeaSE-RN50-PDARTS (ours) &\textbf{15.13}$\pm$0.07 &3.6 &1.0\\
        \bottomrule
    \end{tabular}
    \label{tab:cifar100}
\end{table}

\begin{table}[t]
\caption{
    Test error on CIFAR-10, number of model weights (millions), and search cost (GPU days on a Tesla v100).  
    Results marked with * are obtained from DARTS$^{-}$ \citep{abs-2009-01027}, NoisyDARTS \citep{abs-2005-03566},  and DrNAS \citep{abs-2006-10355}. The rest notations are the same as those in Table~\ref{tab:cifar100}. 
    }
    \begin{adjustbox}{width=0.73\columnwidth,center}
    \centering
    \begin{tabular}{l|ccc}
    \toprule
    Method& Error(\%)& Param(M) & Cost\\
    \midrule
    *DenseNet
    \citep{HuangLMW17}&3.46&25.6 &-\\
    \hline
     *HierEvol \citep{liu2017hierarchical}&3.75$\pm$0.12& 15.7 &300\\
    *NAONet-WS \citep{LuoTQCL18} & 3.53 & 3.1&0.4 \\
        *PNAS \citep{LiuZNSHLFYHM18} &3.41$\pm$0.09  &3.2& 225\\
        *ENAS \citep{pham2018efficient} &2.89 & 4.6  &0.5 \\
    *NASNet-A \citep{zoph2018learning} & 2.65 & 3.3& 1800\\
    *AmoebaNet-B \citep{real2019regularized} & 2.55$\pm$0.05 & 2.8&3150  \\
    \hline
        *R-DARTS \citep{ZelaESMBH20} &2.95$\pm$0.21  &- & 1.6 \\
            *GDAS \citep{DongY19}&2.93& 3.4& 0.2 \\
        *GTN~\citep{abs-1912-07768}& 2.92$\pm$0.06 & 8.2&  0.67\\
    *SNAS \citep{xie2018snas} &2.85 & 2.8& 1.5\\
                  ${}^{\Delta}$DARTS$^{+}$ \citep{abs-1909-06035}&2.83$\pm$0.05&3.7&0.4\\
        *BayesNAS \citep{ZhouYWP19} &2.81$\pm$0.04 &3.4&0.2 \\
        *MergeNAS \citep{WangXYYHS20} &2.73$\pm$0.02 &2.9 & 0.2 \\
        *NoisyDARTS \citep{abs-2005-03566} &2.70$\pm$0.23&3.3  & 0.4 \\
            *ASAP \citep{NoyNRZDFGZ20} &2.68$\pm$0.11 & 2.5&0.2 \\
                *SDARTS
    \citep{abs-2002-05283}&2.61$\pm$0.02 & 3.3& 1.3 \\
                 *DARTS$^{-}$ \citep{abs-2009-01027}&2.59$\pm$0.08&  3.5&0.4\\
             ${}^{\dag}$DARTS$^{-}$ \citep{abs-2009-01027}& 2.97$\pm$0.04& 3.3&0.4\\
            *DropNAS \citep{HongL0TWL020} &2.58$\pm$0.14 & 4.1&0.6 \\
            *PC-DARTS \citep{abs-1907-05737} &2.57$\pm$0.07&3.6& 0.1\\
    *FairDARTS \citep{abs-1911-12126} &2.54 &3.3 &0.4 \\
       *DrNAS \citep{abs-2006-10355} &2.54$\pm$0.03&4.0&  0.4\\
    \hline
        \hline
            *DARTS-1st \citep{liu2018darts} &3.00$\pm$0.14&3.3&  0.4\\
        $\;\;$LeaSE-R18-DARTS1st (ours) &2.85$\pm$0.09 &3.4&0.6\\
              $\;\;$LeaSE-R50-DARTS1st (ours) &\textbf{2.76}$\pm$0.03 &3.3&0.7 \\
        \hline
               *DARTS-2nd \citep{liu2018darts} &2.76$\pm$0.09&3.3&  1.5\\
           $\;\;$LeaSE-R18-DARTS2nd (ours)  & 2.59$\pm$0.06 &3.3 & 1.5 \\
            $\;\;$LeaSE-R50-DARTS2nd (ours) &   \textbf{2.52}$\pm$0.04  &3.4 & 1.7 \\
     \hline
       *PC-DARTS~\citep{abs-1907-05737}& 2.57$\pm$0.07 & 3.6 & 0.1 \\
        $\;\;$LeaSE-R18-PC-DARTS (ours) & 2.50$\pm$0.04 &3.7 &0.4 \\
        $\;\;$LeaSE-R50-PC-DARTS (ours) & \textbf{2.48}$\pm$0.02 &3.7 &0.5 \\
                       \hline
    *P-DARTS \citep{chen2019progressive}& 2.50&3.4&  0.3\\
     $\;\;$LeaSE-R18-PDARTS (ours)& 2.45$\pm$0.03 & 3.4 & 0.8 \\
          $\;\;$LeaSE-R50-PDARTS (ours)& \textbf{2.44}$\pm$0.03 & 3.4 & 1.0\\
        \bottomrule
    \end{tabular}
    \end{adjustbox}
    \label{tab:cifar10}
\end{table}

\begin{table*}[t]
\caption{Top-1 and top-5 classification errors on the test set of ImageNet, number of model parameters (millions) and search cost (GPU days). Results marked with * are obtained from DARTS$^{-}$ \citep{abs-2009-01027} and DrNAS \citep{abs-2006-10355}. The rest notations are the same as those in Table~\ref{tab:cifar100}. From top to bottom, on the first three blocks are 1) networks manually designed by humans; 2) non-gradient based NAS methods; and 3) gradient-based NAS methods. 
    }
    \centering
        \begin{adjustbox}{width=0.98\columnwidth,center}
    \begin{tabular}{l|cccc}
    \toprule
  \multirow{ 2}{*}{Method}   & Top-1  &Top-5 &Param & Cost \\
         & Error (\%) & Error (\%)&(M) & (GPU days)\\
    \midrule
    *Inception-v1 \citep{googlenet}&30.2 &10.1&6.6&- \\
    *MobileNet \citep{HowardZCKWWAA17} &  29.4& 10.5 &4.2&- \\
    *ShuffleNet 2$\times$ (v1) \citep{ZhangZLS18} &  26.4 &10.2 & 5.4&-\\
    *ShuffleNet 2$\times$ (v2) \citep{MaZZS18} &  25.1 &7.6 & 7.4&-\\
    \hline
    *NASNet-A \citep{zoph2018learning} &26.0 &8.4 &5.3 &1800 \\
    *PNAS \citep{LiuZNSHLFYHM18} &25.8 &8.1  &5.1 &225 \\
    *MnasNet-92 \citep{TanCPVSHL19} & 25.2 & 8.0& 4.4&1667\\
        *AmoebaNet-C \citep{real2019regularized} &  24.3 &7.6 &6.4&3150 \\
    \hline
     *SNAS-CIFAR10 \citep{xie2018snas} & 27.3 &9.2 &4.3 &1.5 \\
          *BayesNAS-CIFAR10 \citep{ZhouYWP19} &26.5 &8.9 &3.9&0.2 \\
                    *PARSEC-CIFAR10 \citep{abs-1902-05116} & 26.0 &8.4&5.6&1.0 \\
     *GDAS-CIFAR10 \citep{DongY19} &  26.0&8.5 &5.3 & 0.2\\
                 *DSNAS-ImageNet \citep{HuXZLSLL20} &25.7& 8.1 &- & -\\
          *SDARTS-ADV-CIFAR10 \citep{abs-2002-05283}&25.2& 7.8 &5.4& 1.3 \\
           *PC-DARTS-CIFAR10 \citep{abs-1907-05737} & 25.1 &7.8&5.3&0.1\\
                *ProxylessNAS-ImageNet \citep{cai2018proxylessnas} & 24.9 &7.5 &7.1 &8.3  \\
          *FairDARTS-CIFAR10 \citep{abs-1911-12126} &24.9 &7.5 &4.8 &0.4 \\
     *FairDARTS-ImageNet \citep{abs-1911-12126} &24.4 &7.4 &4.3 &3.0 \\
             *DrNAS-ImageNet \citep{abs-2006-10355} & 24.2 &7.3& 5.2&3.9\\
         *DARTS$^{+}$-ImageNet \citep{abs-1909-06035}& 23.9& 7.4&5.1&6.8\\
        *DARTS$^{-}$-ImageNet \citep{abs-2009-01027}&23.8& 7.0&4.9&4.5\\
     *DARTS$^{+}$-CIFAR100 \citep{abs-1909-06035}&23.7& 7.2&5.1&0.2\\
     \hline
       \hline
       *DARTS2nd-CIFAR10 \citep{liu2018darts}  & 26.7 &8.7&4.7&1.5 \\
        $\;\;$LeaSE-R18-DARTS2nd-CIFAR10 (ours) & \textbf{24.7} & \textbf{8.3}&4.8 &1.5 \\
        \hline
          *PDARTS (CIFAR10) \citep{chen2019progressive}&24.4 &7.4&4.9&0.3\\
        $\;\;$LeaSE-R18-PDARTS-CIFAR10 (ours) & \textbf{23.8} & \textbf{6.7} & 5.0& 0.8 \\
        \hline
             *PDARTS (CIFAR100) \citep{chen2019progressive}&24.7& 7.5&5.1&0.3\\
           $\;\;$LeaSE-R18-PDARTS-CIFAR100 (ours)&\textbf{23.9} & \textbf{6.7}   &5.1& 0.8\\
          \hline
            *PCDARTS-ImageNet \citep{abs-1907-05737} &  24.2 &7.3&5.3&3.8\\
              $\;\;$LeaSE-R18-PCDARTS-ImageNet (ours)& \textbf{22.1} & \textbf{6.0} & 5.5&4.0 \\
        \bottomrule
    \end{tabular}
    \end{adjustbox}
    \label{tab:imagenet}
\end{table*}

Table~\ref{tab:cifar100} shows the results on CIFAR-100, including classification errors on the test set, number of model parameters, and search cost. By comparing different methods, we make the following observations. \textbf{First}, applying LeaSE to different NAS methods, including DARTS, P-DARTS, and PC-DARTS, the classification errors of these methods are greatly reduced. For example, the original error of DARTS-2nd is 20.58\%; when LeaSE is applied, this error is significantly reduced to 16.39\%. As another example, after applying LeaSE to PC-DARTS, the error is reduced from 17.96\% to 16.17\%. Similarly, with the help of LeaSE, the error of P-DARTS is decreased from 17.49\% to 15.13\%. These results strongly demonstrate the broad effectiveness of our framework in searching better neural architectures. The reason behind this is: in our framework, the explanations made by the explainer are used to train the audience model; the validation performance of the audience reflects how good the explanations are; to make good explanations, the explainer's model has to be trained well; driven by the goal of helping the audience to learn well, the explainer continuously improves the training of itself. Such an explanation-driven learning mechanism is lacking in baseline methods, which are hence inferior to our method. 
\textbf{Second}, an audience model with a more expressive architecture can help the explainer to learn better. We experimented with two architectures for the audience model: ResNet with 18 layers (RN18) and ResNet with 50 layers (RN50), where RN50 is more expressive than RN18 since it has more layers. As can be seen, in LeaSE applied to DARTS, PC-DARTS, and P-DARTS, using RN50 as the audience achieves better performance than using RN18. For example, LeaSE-R50-DARTS2nd achieves an error of 16.39\%, which is lower than the 16.80\% error of LeaSE-R18-DARTS2nd. When replacing the audience's architecture from RN18 to RN50, the error of LeaSE-DARTS1st is reduced from 17.04\% to 16.87\%, the error  of LeaSE-PCDARTS is reduced from 16.39\% to 16.17\%, and the error of LeaSE-PDARTS is reduced from 15.23\% to 15.13\%.   The reason is that to help a stronger audience to learn better, the explainer has to be even stronger. For a stronger audience model, it already has great capability in achieving excellent classification performance. To further improve this audience, the explanations used to train this audience need to be very sensible and informative. To generate such  explanations, the explainer has to force itself to learn very well. 
\textbf{Third}, our LeaSE-RN50-PDARTS method achieves the lowest error among all methods listed in this table, which indicates that our method is very competitive in driving the field of NAS research to a new state-of-the-art. \textbf{Fourth}, the performance gain of our method does not come at a cost of substantially increasing model size and search cost: the number of model parameters in architectures searched by our methods are at a similar level compared with those by other methods; so are the search costs.

In Table~\ref{tab:cifar10}, we show the results on CIFAR-10, including classification errors on the test set, number of model parameters, and search cost. The observations made from this table are similar to those from Table~\ref{tab:cifar100}. \textbf{First}, with the help of our LeaSE framework, the classification errors of DARTS, PC-DARTS, and P-DARTS are all reduced. For example, applying LeaSE to DARTS-2nd manages to reduce the error of DARTS-2nd from 2.76\% to 2.52\%. As another example, applying LeaSE to P-DARTS decreases the error from 2.50\% to 2.44\%. This further demonstrates the effectiveness of explanation-driven learning. 
\textbf{Second}, an audience with a stronger architecture helps the explainer to learn better. For example, in LeaSE-DARTS, LeaSE-PDARTS, and LeaSE-PCDARTS, when the audience is set to RN50, the performance is better, compared with setting the audience to RN18. \textbf{Third}, our method LeaSE-R50-PDARTS achieves the lowest error among all methods in this table, which further demonstrates its great potential in continuously pushing the limit of NAS research. \textbf{Fourth}, while the architectures searched by our framework yield better performance, their model size and search cost are not substantially increased compared with baselines.

In Table~\ref{tab:imagenet}, we compare different methods on ImageNet, in terms of top-1 and top-5 classification errors on the test set, number of model parameters, and search cost. In experiments based on PC-DARTS, the architectures are searched on a subset of ImageNet. In other experiments, the architectures are searched on CIFAR-10 and CIFAR-100. LeaSE-R18-DARTS2nd-CIFAR10 denotes that LeaSE is applied to DARTS-2nd and performs search on CIFAR10, with the audience model set to ResNet-18. Similar meanings hold for other notations in such a format. 
The observations made from these results are consistent with those made from Table~\ref{tab:cifar100} and Table~\ref{tab:cifar10}. The architectures searched using our methods are consistently better than those searched by corresponding baselines. For example, LeaSE-R18-DARTS2nd-CIFAR10 achieves lower top-1 and top-5 errors than DARTS2nd-CIFAR10. LeaSE-R18-PDARTS-CIFAR10 outperforms PDARTS-CIFAR10. These results again show that by explaining well, a model can gain better predictive performance. 
The model size and search cost of architectures searched by our methods are on par with those in other methods, which demonstrates that the performance gain of our framework is obtained without sacrificing compactness of architectures or computational efficiency substantially. Among all the methods in this table, our method LeaSE-R18-PCDARTS-ImageNet achieves the lowest top-1 and top-5 errors, which further demonstrates the great effectiveness of our method.

\subsection{Ablation Studies}
In this section, we perform ablation studies to investigate the importance of individual components in our framework. In each ablation study, we compare the ablation setting with the full framework. Specifically, we study the following ablation settings. 
\begin{itemize}[leftmargin=*]
    \item \textbf{Ablation setting 1}. In this setting, the explainer updates its architecture by minimizing the validation loss of the audience only, without considering the validation loss of itself. 
     The corresponding formulation is: 
 \begin{equation*}
\begin{array}{l}
\underset{A}{\textrm{min}}
  \;\;  
    \;  L(W^*( \Delta^*(E^*(A)) ), D_a^{(\textrm{val})})\\
      s.t. \; W^*( \Delta^*(E^*(A)) ) =
      \underset{W}{\textrm{min}}
      \;\; \sum\limits_{i=1}^N \ell(f(\delta_i^*(E^*(A))\odot x_i;W),t_i)\\
    \quad \Delta^*(E^*(A)) =
    \underset{\Delta}{\textrm{max}}
    \;\; \sum\limits_{i=1}^N \ell(f(x_i+\delta_i;{E^*(A)}),f(x_i;{E^*(A)})) \\
     \quad   E^*(A) =
     \underset{E}{\textrm{min}}
     \; L(E,A,D_e^{(\textrm{tr})}). 
\end{array}
\end{equation*}
During this study, we  set the architecture of the audience to ResNet-18. On  CIFAR-100, LeaSE is applied to P-DARTS. On CIFAR-10, LeaSE is applied to DARTS-2nd. 
    \item \textbf{Ablation study on $\gamma$}. We investigate how the tradeoff parameter $\gamma$ in Eq.(\ref{eq:lease}) affects the classification errors of the explainer. For both CIFAR-100 and CIFAR-10, 5K images are uniformly sampled from the 50K training and validation examples. The 5K images are used as a test set for reporting the architecture evaluation performance, where the architecture is searched on the rest 45K images. We applied LeaSE to P-DARTS and chose ResNe-18 as the architecture of the audience's model. 
\end{itemize}

\begin{table}[t]
\caption{Results for ablation setting 1. “Audience only” means that only the audience's validation loss is minimized to update the architecture of the explainer. “Audience + explainer” means that both the validation loss of the audience and the validation loss of the explainer itself are minimized to learn the explainer's architecture. 
    }
    \centering
    \begin{tabular}{l|c}
    \hline
    Method & Error (\%)\\
    \hline
    Audience  only (CIFAR-100) &  16.08$\pm$0.15 \\ 
            Audience + explainer  (CIFAR-100) &\textbf{15.23}$\pm$0.11  \\
         \hline
             Audience only (CIFAR-10) & 2.72$\pm$0.07  \\
         Audience  + explainer  (CIFAR-10) &\textbf{2.59}$\pm$0.06   \\
         \hline
    \end{tabular}
    \label{tab:ab1}
\end{table}

In Table~\ref{tab:ab1}, we show the results on CIFAR-10 and CIFAR-100 under the ablation setting 1. On both datasets, ``audience + explainer" where the validation losses of both the audience model and explainer itself are minimized to update the explainer's architecture works better than ``audience only" where only the audience's validation loss is used to learn the architecture. Audience's validation loss reflects how good the explanations made by the explainer are. Explainer's validation loss reflects how strong the explainer's prediction ability is. Combining these two losses provides more useful feedback to the explainer than using one loss only, which hence can help the explainer to learn better.

\begin{figure}[t]
    \centering
 \includegraphics[width=0.49\columnwidth]{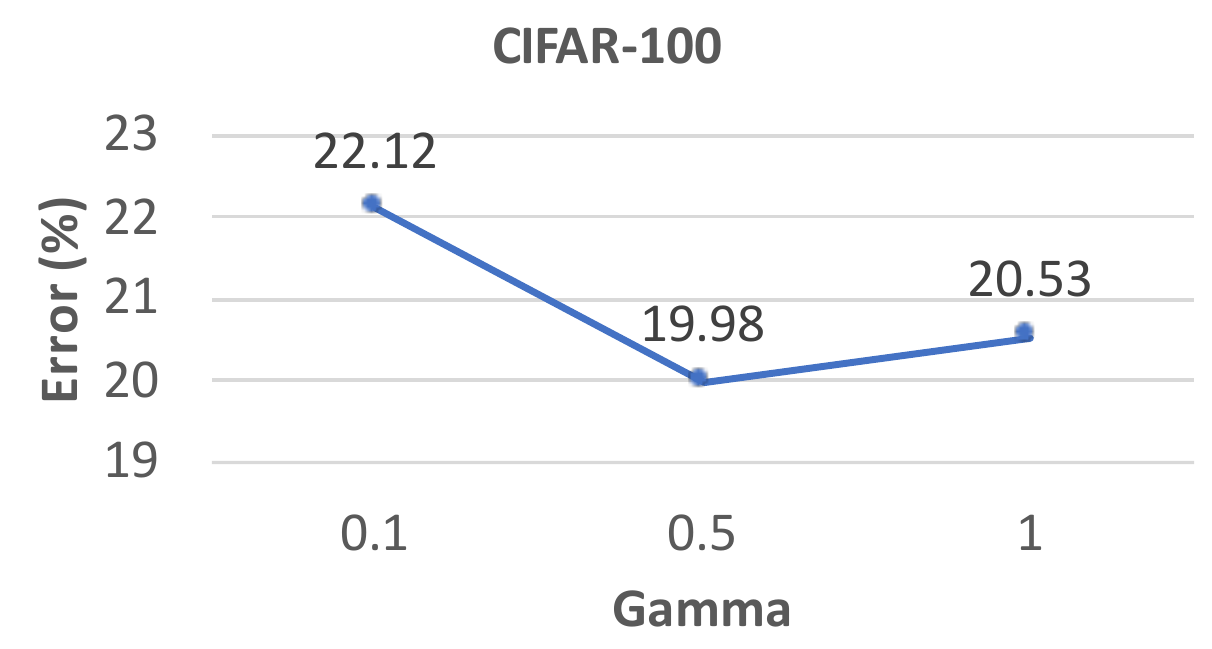}
  \includegraphics[width=0.49\columnwidth]{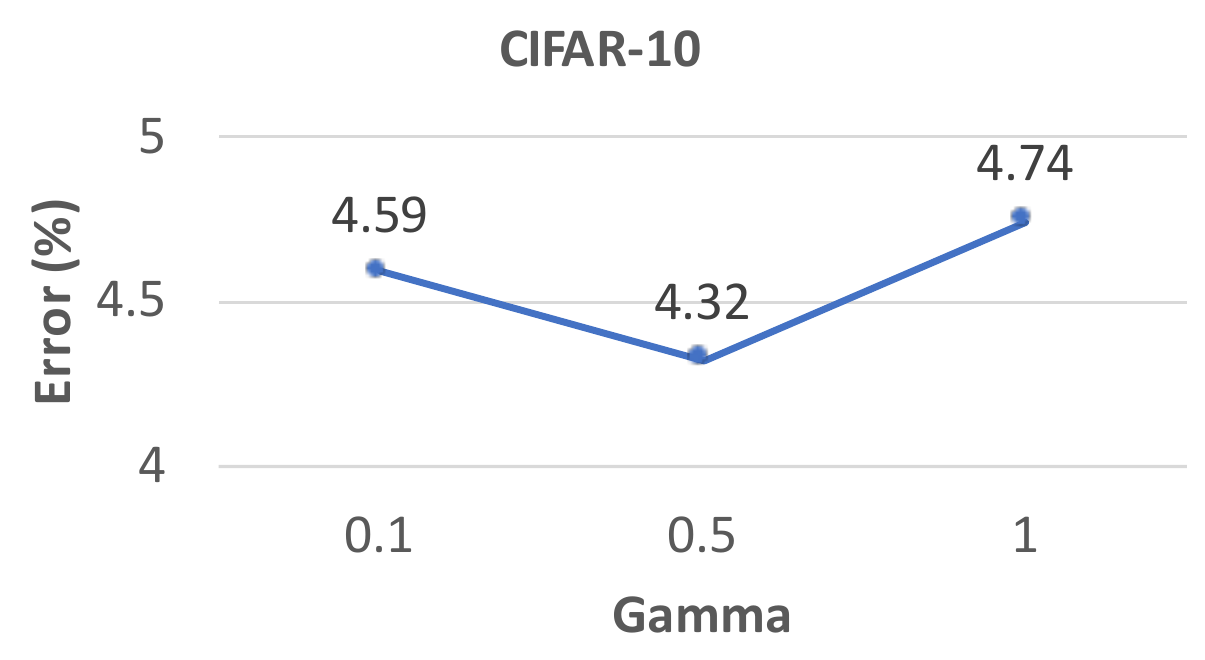}
       \caption{How errors change as $\gamma$ increases.}
 \label{fig:gamma}
\end{figure}

In Figure~\ref{fig:gamma}, we show how LeaSE's classification errors on the test sets of CIFAR-10 and CIFAR-100 vary as we increase the tradeoff parameter $\gamma$. The curve on CIFAR-100 shows that the error decreases when we increase $\gamma$ from 0.1 to 0.5. The reason is that a larger $\gamma$ enables the audience to provide stronger feedback to the explainer regarding how good the explanations are. Such feedback can guide the explainer to refine its architecture for generating better explanations.
However, if $\gamma$ is further increased, the error becomes worse. Under such circumstances, too much emphasis is put on evaluating how good the explanations are and less attention is paid to the predictive ability of the explainer. The architecture is biased to generating good explanations with predictive performance compromised, which leads to inferior performance. 
A similar trend is shown in the curve on CIFAR-10.

\section{Conclusions}
Motivated by humans' explanation-driven learning skill, we develop a novel machine learning framework referred to as learning by self-explanation (LeaSE). 
In LeaSE, the primary goal is to help an explainer model learn how to well perform a target task. The way to achieve this goal is to let the explainer make sensible explanations. The intuition behind LeaSE is that a model has to learn to understand a topic very well before it can explain this topic  clearly. A four-level optimization framework is developed to formalize LeaSE, where the learning is organized into four  stages: the explainer learns a topic; the explainer explains this topic; the audience learns this topic based on the explanations given by the explainer; the explainer re-learns this topic based on  the learning outcome of the audience. We apply LeaSE for  neural architecture search on image classification datasets including CIFAR-100, CIFAR-10, and ImageNet. Experimental results strongly demonstrate the effectiveness of our proposed method.

\bibliography{release-2}

\end{document}